# A Fuzzy Brute Force Matching Method for Binary Image Features

Erkan Bostanci[1], Nadia Kanwal[2] Betul Bostanci[3] and Mehmet Serdar Guzel[1]

[1](Computer Engineering Department, Ankara University, Turkey
{ebostanci, mguzel}@ankara.edu.tr)
2(Lahore College for Women University, Lahore, Pakistan)
nadia.kanwal@lcwu.edu.pk)
3(HAVELSAN Inc., Turkey
bbostanci@havelsan.com.tr)

***Abstract:*** *-* Matching of binary image features is an important step in many different computer vision applications. Conventionally, an arbitrary threshold is used to identify a correct match from incorrect matches using Hamming distance which may improve or degrade the matching results for different input images. This is mainly due to the image content which is affected by the scene, lighting and imaging conditions. This paper presents a fuzzy logic based approach for brute force matching of image features to overcome this situation. The method was tested using a well-known image database with known ground truth. The approach is shown to produce a higher number of correct matches when compared against constant distance thresholds. The nature of fuzzy logic which allows the vagueness of information and tolerance to errors has been successfully exploited in an image processing context. The uncertainty arising from the imaging conditions has been overcome with the use of compact fuzzy matching membership functions.

***Keywords:*** *- fuzzy matching, Hamming distance, image features*

## I. INTRODUCTION

Finding correspondences between two or more views of the same scene is the primary step in many vision algorithms including tracking or reconstruction. These correspondences are found by first detecting prominent regions, i.e. features, of the scene content such as the corners or blobs. The next step is describing the features with descriptors using local pixel information around them. These descriptors are compared with each other in order to find feature correspondences between images, a process known as feature matching [1,2] as shown in Figure 1.

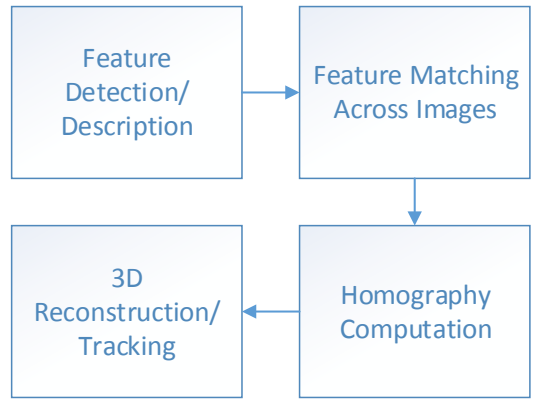

**Figure 1 Process Diagram**

Recent research has presented a number of binary feature detectors such as BRIEF [3], ORB [4] or BRISK [5]. There are two main reasons behind this interest in binary descriptors. First, they have a compact representation. Second, they can be matched efficiently using options provided by modern processor instructions sets with the Hamming distance [6].

As will be mentioned in the next section, conventional use of threshold may create problems such as inconsistent number of matching features each time the threshold value is changed. This, in due course, results in further problems when homographies are to be computed based on these feature correspondences.

This paper presents the use of fuzzy logic in order to compensate the problems related to a static use of threshold values. Based on a compact rule-base the approach can yield more stable feature numbers.

The rest of the paper is structured as follows: Section 2 describes the background on conventional use of threshold followed by Section 3 where fuzzy matching approach is presented. Section 4 presents the matching results with 3 static thresholds and the fuzzy approach. Finally, the paper is concluded in Section 5.

## II. CONVENTIONAL USE OF THRESHOLD

As a distance metric the Hamming distance can be defined based on the following quantities:

```
f00=num. of positions where both descriptors have 0s.
f01=num. of positions where the first has 0 and the second has a 1.
f10=num. of positions where the first has 1 and the second has a 0.
f11=num. of positions where both descriptors have 1s.
```

Based on these quantities, the Hamming distance $d_H$ is defined as follows:

```
dH=(f11+f00)/(f00+f01+f10+f11)
```

which can be used along with a constant threshold (*t*), for instance, if $d_H < t$, then a match is found. Using this distance, two views of the same scene can be matched.

A matching algorithm compares feature descriptors to find a match. The above mentioned method of comparing against a threshold can produce false matches by identifying regions with similarities at different or multiple locations. The reason behind this is that it is quite possible that an image had different regions having a similar texture. For this reason, it is also required to check whether these matches are correct. This is performed by using an approach called Random Sampling Consensus (RANSAC) to find an estimate of the homography matrix which represents the perspective transformation between two images.

This estimate is certainly affected by the matches, resulting in inaccurate estimations when the majority of the matches are incorrect. In some datasets, such as the one used here (available at http://www.robots.ox.ac.uk/~vgg/data/data-aff.htm), the ground-truth homography matrices for the datasets are available for verification.

## III. FUZZY THRESHOLDING

Instead of checking against a constant distance threshold, we propose a simple fuzzy method for brute force matching in which all descriptors describing the features in the first image are compared against the ones in the second image.

For this approach, the following input and output membership functions given in Figures 1 and 2 are defined.

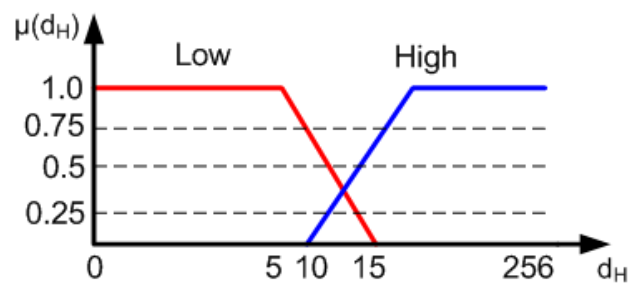
**Figure 2 Input membership function**

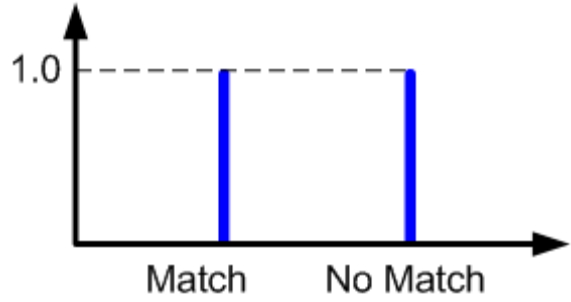

**Figure 3 Output membership function**

These membership functions are employed to operate on a very compact rule base:

```
If Hamming distance d_H is LOW then the features MATCH
If Hamming d_H is HIGH then the features DO NOT MATCH
```

During the matching process, the distance between the descriptors for two image features is computed with Hamming distance. Then this distance is used as the crisp input of the fuzzy system. The membership values of this crisp value are found for two fuzzy set Low and High. The rules are evaluated and finally the output decision is obtained from the zero-order Sugeno type output membership function (singleton) as a `Match` or `No Match`.

## IV. RESULTS

Results for the fuzzy approach for identifying the matches are presented in Table 1.

**Table 1 Matching comparison for constant and fuzzy thresholding**

| Dataset | t=5 | | t=10 | | t=15 | | Fuzzy | |
|---|---|---|---|---|---|---|---|---|
| | *M* | *CM* | *M* | *CM* | *M* | *CM* | *M* | *CM* |
| Bark (1-2) | *12* | *6* | *169* | *17* | *1220* | *17* | *642* | *25* |
| Bark (1-3) | *4* | *2* | *152* | *8* | *1207* | *13* | *635* | *10* |
| Bark (1-4) | *4* | *3* | *139* | *4* | *1151* | *9* | *592* | *6* |
| Bark (1-5) | *1* | *1* | *139* | *2* | *1205* | *3* | *656* | *3* |
| Bark (1-6) | *70* | *1* | *290* | *1* | *1300* | *1* | *774* | *1* |
| Bikes (1-2) | 51 | 47 | 327 | 101 | 1587 | 147 | 998 | 134 |
| Bikes (1-3) | 48 | 40 | 401 | 106 | 1652 | 145 | 1056 | 132 |
| Bikes (1-4) | 41 | 28 | 386 | 90 | 1575 | 122 | 1034 | 115 |
| Bikes (1-5) | 43 | 26 | 483 | 81 | 1632 | 102 | 1169 | 98 |
| Bikes (1-6) | 39 | 19 | 490 | 57 | 1609 | 80 | 1123 | 75 |
| Boat (1-2) | 41 | 36 | 248 | 77 | 1213 | 103 | 733 | 95 |
| Boat (1-3) | 29 | 25 | 219 | 66 | 1253 | 97 | 722 | 89 |
| Boat (1-4) | 13 | 10 | 215 | 37 | 1294 | 58 | 740 | 52 |
| Boat (1-5) | 5 | 4 | 148 | 13 | 1166 | 28 | 604 | 22 |
| Boat (1-6) | 2 | 1 | 141 | 9 | 1179 | 30 | 655 | 22 |
| Graffiti (1-2) | 15 | 13 | 185 | 49 | 1219 | 81 | 674 | 68 |
| Graffiti (1-3) | 5 | 2 | 166 | 12 | 1165 | 34 | 639 | 27 |
| Graffiti (1-4) | 1 | 1 | 137 | 6 | 1072 | 17 | 554 | 13 |
| Graffiti (1-5) | 1 | 1 | 128 | 1 | 1087 | 3 | 553 | 2 |
| Graffiti (1-6) | 1 | 1 | 129 | 1 | 1142 | 1 | 581 | 1 |
| Leuven (1-2) | 35 | 25 | 222 | 55 | 1264 | 89 | 709 | 75 |

| | | | | | | | |
|---|---|---|---|---|---|---|---|
| Leuven (1-3) | 41 | 27 | 236 | 47 | 1264 | 78 | 698 | 63 |
| Leuven (1-4) | 11 | 6 | 168 | 26 | 1213 | 59 | 640 | 44 |
| Leuven (1-5) | 8 | 2 | 165 | 16 | 1159 | 33 | 644 | 25 |
| Leuven (1-6) | 8 | 4 | 147 | 16 | 1147 | 39 | 609 | 25 |
| Trees (1-2) | 25 | 21 | 194 | 40 | 1262 | 59 | 699 | 53 |
| Trees (1-3) | 26 | 22 | 183 | 40 | 1279 | 57 | 690 | 53 |
| Trees (1-4) | 16 | 14 | 183 | 28 | 1235 | 40 | 707 | 38 |
| Trees (1-5) | 7 | 4 | 161 | 11 | 1276 | 23 | 680 | 19 |
| Trees (1-6) | 4 | 2 | 131 | 8 | 1208 | 11 | 633 | 8 |
| UBC (1-2) | 47 | 47 | 217 | 72 | 1298 | 87 | 735 | 81 |
| UBC (1-3) | 35 | 35 | 216 | 58 | 1265 | 76 | 734 | 68 |
| UBC (1-4) | 30 | 29 | 195 | 57 | 1242 | 74 | 677 | 71 |
| UBC (1-5) | 21 | 18 | 192 | 49 | 1289 | 62 | 687 | 58 |
| UBC (1-6) | 16 | 15 | 152 | 30 | 1181 | 45 | 631 | 42 |
| Wall (1-2) | 4 | 4 | 117 | 17 | 1092 | 46 | 559 | 35 |
| Wall (1-3) | 5 | 4 | 105 | 18 | 1071 | 56 | 520 | 34 |
| Wall (1-4) | 2 | 1 | 91 | 3 | 1123 | 20 | 560 | 12 |
| Wall (1-5) | 1 | 1 | 108 | 2 | 1143 | 11 | 546 | 9 |
| Wall (1-6) | 1 | 1 | 88 | 1 | 1168 | 1 | 549 | 1 |

From the results, the effect of changing the threshold value can be seen. For the first image pair (Bark 1-2), when the threshold is 5 matching yields 12 matches out of which there are only 6 matches. A threshold of 10 gives 169 with only 17 correct matches. When `t=15`, even more matches are identified but still not much improvement in the number of correct matches. The fuzzy approach presented in the paper, performs a better automatic selection of the threshold since the image content, hence the result of the feature descriptor, are susceptible to the changes in the imaging conditions.

Looking at the results, it can be confidently said that the fuzzy approach can handle the matching process much better than when different thresholds are solely used. The input membership function of Figure 2 covers all the three different values of the thresholds. The initial matches shown with M are closer to the ones confirmed by the homography (CM) when the fuzzy method is used as compared to when the threshold is selected as 15. Most of the results produce a higher `CM/M` ratio indicating that the matches found using the fuzzy approach are confirmed by the homography result as well.

The fuzzy decision making for a match takes only 0.003731 milliseconds per pairwise feature match, still under a millisecond for a hundred feature pairs.

## V. CONCLUSION

This paper proposed a fuzzy approach for brute force matching of binary image features without specifying constant thresholds which definitely affect the number of initial matches and the ones confirmed by the homography estimation. The results of this adaptive approach are promising in terms of showing the power of fuzzy logic in the presence of uncertainty in the matching process i.e. how to select the threshold for different datasets which are subject to affine and photometric transformations.

Future work will investigate how to integrate this approach with the feature description process again using a fuzzy logic framework for better modelling of the feature representations in varying imaging conditions.